\documentclass[11pt]{article}
\usepackage{acl2012}
\usepackage{times}
\usepackage{latexsym}
\usepackage{amsmath}
\usepackage{graphicx}
\usepackage{xcolor}
\usepackage{amsfonts}
\usepackage{amsmath}
\usepackage{url}
\usepackage{graphicx}
\usepackage{array}
\usepackage{multicol}

\makeatletter
\newcommand{\@BIBLABEL}{\@emptybiblabel}
\newcommand{\@emptybiblabel}[1]{}
\makeatother
\usepackage{hyperref}

\newcommand{\model}[1]{\textsc{#1}}
\usepackage{multirow}

\newcolumntype{P}[1]{>{\raggedright\arraybackslash}p{#1}}

\usepackage{dcolumn}

\title{From Visual Attributes to Adjectives\\through 
  Decompositional Distributional Semantics}
\author{Angeliki Lazaridou~~~Georgiana Dinu\thanks{~~Current affiliation: Thomas J. Watson Research Center, IBM, gdinu@us.ibm.com}~~~Adam Liska~~~Marco Baroni\\
  Center for Mind/Brain Sciences\\
  University of Trento\\
  {\tt \{angeliki.lazaridou|georgiana.dinu|adam.liska|marco.baroni\}@unitn.it} \\}

\date{}

\begin{document}
\maketitle

\begin{abstract}
  As automated image analysis progresses, there is increasing interest
  in richer linguistic annotation of pictures, with \emph{attributes}
  of objects (e.g., \emph{furry}, \emph{brown}\ldots) attracting most
  attention. By building on the recent ``zero-shot learning''
  approach, and paying attention to the linguistic nature of
  attributes as noun modifiers, and specifically adjectives, we show
  that it is possible to tag images with attribute-denoting adjectives
  even when no training data containing the relevant annotation are
  available. Our approach relies on two key observations. First,
  objects can be seen as bundles of attributes, typically expressed as
  adjectival modifiers (a \emph{dog} is something \emph{furry},
  \emph{brown}, etc.), and thus a function trained to map visual
  representations of objects to nominal labels can implicitly learn to
  map attributes to adjectives. Second, objects and attributes come
  together in pictures (the same thing is a \emph{dog} and it is
  \emph{brown}). We can thus achieve better attribute (and object)
  label retrieval by treating images as ``visual phrases'', and
  decomposing their linguistic representation into an
  attribute-denoting adjective and an object-denoting noun. Our
  approach performs comparably to a method exploiting manual attribute
  annotation, it outperforms various competitive alternatives in both
  attribute and object annotation, and it automatically constructs
  attribute-centric representations that significantly improve
  performance in supervised object recognition.
\end{abstract}

\section{Introduction} 
\label{sec:intro}

As the quality of image analysis algorithms improves, there is
increasing interest in annotating images with linguistic descriptions
ranging from single words describing the depicted objects and their
properties~\cite{Fahradi:etal:2009,Lampert:etal:2009} to richer
expressions such as full-fledged image
captions~\cite{Kulkarni:etal:2011,Mitchell:etal:2012}. This trend has
generated wide interest in linguistic annotations beyond concrete
nouns, with the role of adjectives in image descriptions receiving, in
particular, much attention. %

Adjectives are of special interest because of their central role in
so-called \textit{attribute-centric} image representations. This
framework views objects as bundles of properties, or
\emph{attributes}, commonly expressed by adjectives (e.g.,
\textit{furry}, \textit{brown}), and uses the latter as features to
learn higher-level, semantically richer representations of
objects~\cite{Fahradi:etal:2009}.\footnote{In this paper, we assume
  that, just like nouns are the linguistic counterpart of visual
  objects, visual attributes are expressed by adjectives. An informal
  survey of the relevant literature suggests that, when attributes
  have linguistic labels, they are indeed mostly expressed by
  adjectives. There are some attributes, such as parts, that are more
  naturally expressed by prepositional phrases (PPs: \emph{with a
    tail}).  Interestingly, \newcite{Dinu:Baroni:2014} showed that the
  decomposition function we will adopt here can derive both
  adjective-noun and noun-PP phrases, suggesting that our approach
  could be seamlessly extended to visual attributes expressed by
  noun-modifying PPs.}  Attribute-based methods achieve better
generalization of object classifiers with less training
data~\cite{Lampert:etal:2009}, while at the same time producing
semantic representations of visual concepts that more accurately model
human semantic intuition~\cite{Silberer:etal:2013}. Moreover,
automated attribute annotation can facilitate finer-grained image
retrieval (e.g., searching for a \emph{rocky beach} rather than
a \emph{sandy beach}) and provide the basis for more accurate image
search (for example in cases of \textit{visual sense
  disambiguation}~\cite{Divvala:etal:2014}, where a user disambiguates
their query by searching for images of \textit{wooden cabinet} as
furniture and not just \textit{cabinet}, which can also mean council).

Classic attribute-centric image analysis requires, however, extensive
manual and often domain-specific annotation of
attributes~\cite{Vedaldi:etal:2014}, or, at best, complex unsupervised
image-and-text-mining procedures to learn them~\cite{Berg:etal:2010}.
At the same time, resources with high-quality per-image attribute
annotations are limited; to the best of our knowledge, coverage of all
publicly available datasets containing non-class specific attributes
does not exceed 100 attributes,\footnote{The attribute datasets we are
  aware of are the ones of \newcite{Farhadi:etal:2010},
  \newcite{Ferrari:Zisserman:2007} and
  \newcite{Russakovsky:Fei-Fei:2010}, containing annotations for 64, 7
  and 25 attributes, respectively. (This count excludes the SUN
  Attributes Database \cite{Patterson:etal:2014}, whose attributes
  characterize scenes rather than concrete objects.)}  orders of
magnitude smaller than the equivalent object-annotated datasets
\cite{Deng:etal:2009}.  Moreover, many visual attributes currently
available (e.g., \textit{2D-boxy, furniture leg}), albeit visually
meaningful, do not have straightforward linguistic equivalents,
rendering them inappropriate for applications requiring natural
linguistic expressions, such as the search scenarios considered above.

A promising way to limit manual attribute annotation effort is to
extend recently proposed \emph{zero-shot learning} methods, until now
applied to object recognition, to the task of labeling images with
attribute-denoting adjectives. The zero-shot approach relies on the
possibility to extract, through distributional methods, semantically
effective vector-based word representations from text corpora, on a
large scale and without supervision \cite{Turney:Pantel:2010}. In
zero-shot learning, training images labeled with object names are also
represented as vectors (of features extracted with standard
image-analysis techniques), which are paired with the vectors
representing the corresponding object names in language-based
distributional semantic space. Given such paired training data,
various algorithms
\cite{Socher:etal:2013a,Frome:etal:2013,Lazaridou:etal:2014} can be
used to induce a \emph{cross-modal projection} of images onto
linguistic space.  This projection is then applied to map previously
unseen objects to the corresponding linguistic labels.  The method
takes advantage of the similarities in the vector space topologies of
the two modalities, allowing information propagation from the limited
number of objects seen in training to virtually any object with a
vector-based linguistic representation.

To adapt zero-shot learning to attributes, we rely on their nature as
(salient) properties of objects, and on how this is reflected
linguistically in modifier relations between adjectives and nouns. We
build on the observation that visual and linguistic
attribute-adjective vector spaces exhibit similar structures:
The correlation $\rho$ between the pairwise similarities in
visual and linguistic space of all attributes-adjectives from our
experiments is 0.14 (significant at $p<0.05$).\footnote{In this paper, we report 
significance at  $\alpha=0.05$ threshold.}  
While the
correlation is smaller than for object-noun data (0.23), we conjecture
it is sufficient for zero-shot learning of attributes. We will
confirm this by testing a cross-modal projection function
from attributes, such as colors and shapes, onto adjectives in
linguistic semantic space, trained on pre-existing annotated datasets
covering less than 100 attributes (Experiment 1).

\begin{figure} 
\centering 
\begin{minipage}{.23\textwidth} 
\centering
\includegraphics[scale=0.19]{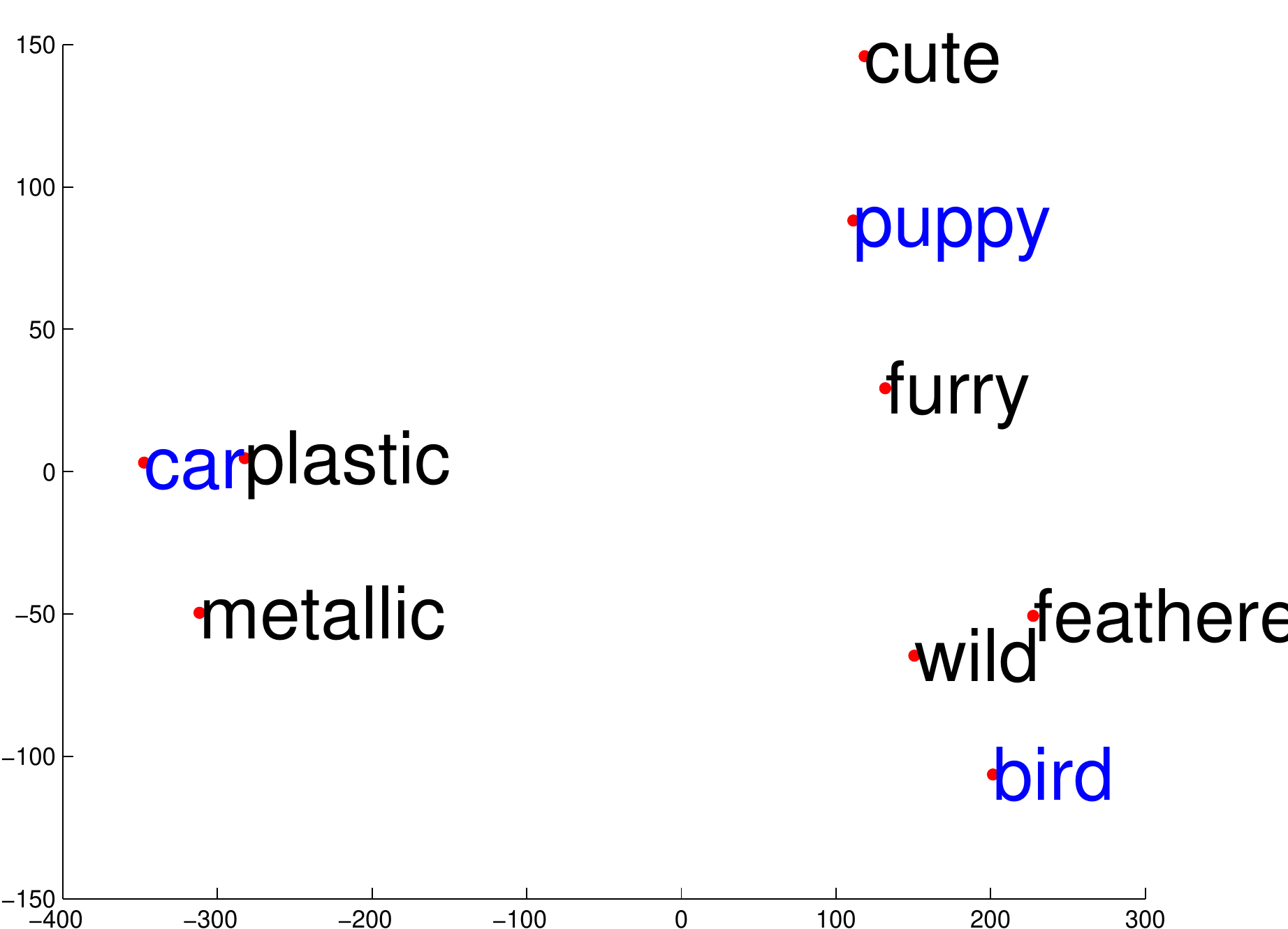} 
\label{pic:vision} 
\end{minipage}
\begin{minipage}{.23\textwidth} 
\centering
\includegraphics[scale=0.19]{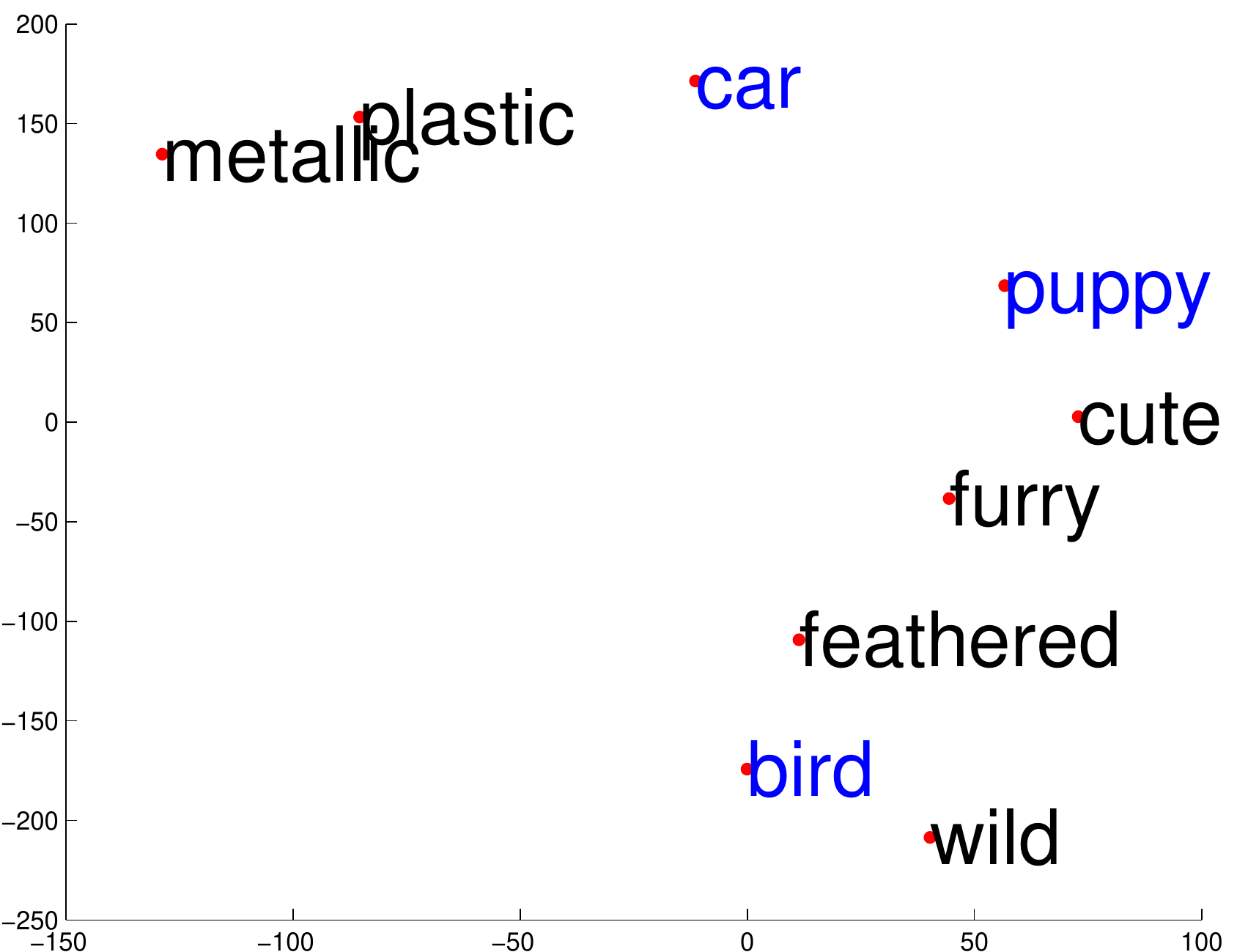} 
\label{pic:language}
\end{minipage}
\caption{t-SNE~\protect\cite{Laurens:Hinton:2008} visualization of 3
objects together with the 2 nearest attributes in our visual
space (left), and of the corresponding nouns and adjectives in 
linguistic space (right).} 
\label{fig:bundle}
\end{figure}

We proceed to develop an approach achieving equally good attribute-labeling
performance \emph{without} manual attribute annotation. Inspired by
linguistic and cognitive theories that characterize objects as attribute
bundles~\cite{Murphy:2002}, we hypothesize that when we learn to project
images of objects to the corresponding noun labels, we implicitly learn to
associate the visual properties/attributes of the objects to the
corresponding adjectives. As an example, Figure~\ref{fig:bundle}
(\textit{left}) displays the nearest attributes of \emph{car}, \emph{bird}
and \emph{puppy} in the visual space and, interestingly,  the relative
distance between the noun denoting objects and the adjective denoting
attributes is also preserved in the linguistic space (\textit{right}).

\begin{figure}
\includegraphics[scale=0.43]{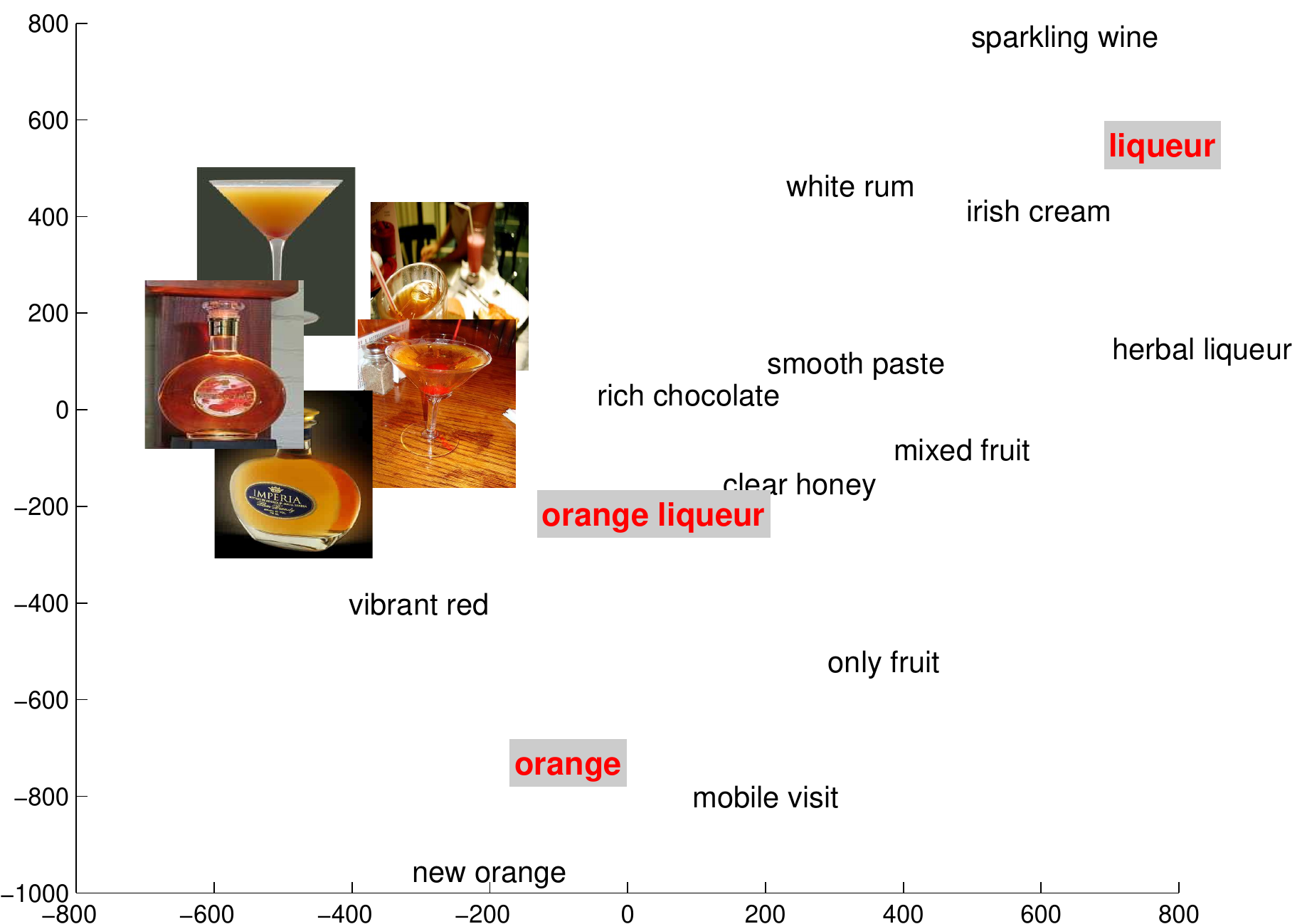}
\caption{Images tagged with \emph{orange} and \emph{liqueur} are
mapped in linguistic space closer to the vector of the phrase
\emph{orange liqueur} than to the \emph{orange} or
\emph{liqueur} vectors (t-SNE visualization) (the figure also
shows the nearest neighbours of phrase, adjective and noun in
linguistic space). The mapping is trained using solely
noun-annotated images.}
\label{fig:visual_phrase} 
\end{figure}

We further observe that, as also highlighted by recent work in object
recognition, any object in an image is, in a sense, a \emph{visual
phrase}~\cite{Sadeghi:Fahradi:2011,Divvala:etal:2014}, i.e., the
object and its attributes are mutually dependent. For example, we
cannot visually isolate the object \textit{drum} from attributes such
as \textit{wooden} and \textit{round}. Indeed, within our data, in
80\% of the cases the projected image of an object is closer to the
semantic representation of a phrase describing it than to either the
object or attribute labels. See Figure \ref{fig:visual_phrase} for an
example.

Motivated by this observation, we turn to recent work in
distributional semantics defining a vector decomposition
framework~\cite{Dinu:Baroni:2014} which, given a vector encoding the
meaning of a phrase, aims at decoupling its constituents, producing
vectors that can then be matched to a sequence of words best capturing
the semantics of the phrase.  We adopt this framework to decompose
image representations projected onto linguistic space into an
adjective-noun phrase. We show that the method yields results
comparable to those obtained when using attribute-labeled training
data, while only requiring object-annotated data. Interestingly, this
\textit{decompositional} approach also doubles the performance of
object/noun annotation over the standard zero-shot approach
(Experiment 2). Given the positive results of our proposed method, we
conclude with an extrinsic evaluation (Experiment 3); we show that
attribute-centric representations of images created with the
decompositional approach boost performance in an object classification
task, supporting claims about its practical utility.

In addition to contributions to image annotation, our work suggests
new test beds for distributional semantic representations of nouns and
associated adjectives, and provides more in-depth evidence of the
potential of the decompositional approach.

\section{General experimental setup}

\subsection{Cross-Modal Mapping}
\label{subsec:mapping}

Our approach relies on \emph{cross-modal mapping} from a visual
semantic space \emph{V}, populated with vector-based representations
of images, onto a linguistic (distributional semantic) space \emph{W}
of word vectors.  The mapping is performed by first inducing a
function $f_{proj}: \mathbb{R}^{d_{1}}\to \mathbb{R}^{d_2}$ from data
points $(v_i, w_i)$, where $v_i \in \mathbb{R}^{d_1}$ is a vector
representation of an image tagged with an object or an attribute (such
as \emph{dog} or \emph{metallic}), and $w_i \in \mathbb{R}^{d_2}$ is
the linguistic vector representation of the corresponding word. The
mapping function can subsequently be applied to any given image $v_i
\in V$ to obtain its projection $w^{\prime}_i \in W$ onto linguistic
space: 
\\[-2ex]
$$w^{\prime}_i = f_{proj}(v_i)$$
Specifically, we consider two mapping methods. In the \model{Ridge}
regression approach, we learn a linear function $F_{proj} \in
\mathbb{R}^{d_2\times d_1}$ by solving the Tikhonov-Phillips
regularization problem, which minimizes the following objective:
$$||W^{Tr} - F_{proj}V^{Tr}||^{2}_{2} - ||\lambda F_{proj}||^{2}_{2},$$ where
$W^{Tr}$ and $V^{Tr}$ are obtained by stacking the word vectors $w_i$
and corresponding image vectors $v_i$, from the training
set.\footnote{The parameter $\lambda$ is determined through
  cross-validation on the training data.}

Second, motivated by the success of Canonical Correlations Analysis
(CCA)~\cite{Hotelling:1936} in several vision-and-language tasks, such
as image and caption retrieval
\cite{Gong:etal:2014,Hardoon:etal:2004,Hodosh:etal:2013}, we adapt
\textit{normalized Canonical Correlations Analysis} (\model{nCCA}) to
our setup. Given two paired observation matrices $X$ and $Y$, in our
case $W^{Tr}$ and $V^{Tr}$, CCA seeks two projection matrices $A$ and
$B$ that maximize the correlation between $A^TX$ and $B^TY$.  This can
be solved efficiently by applying SVD to
\\[-1ex]
$$\footnotesize {\hat{C}_{XX}}^{1/2} {\hat{C}_{XY}}^{ } {\hat{C}_{YY}}^{1/2} =  U\Sigma V^{T}$$
where $\hat{C}$ stands for the covariance matrix.  Finally, the
projection matrices are defined as $A={\hat{C}_{XX}}^{1/2}U$ and
$B={\hat{C}_{YY}}^{1/2}V$. \newcite{Gong:etal:2014} propose a
\emph{normalized} variant of CCA, in which the projection matrices are
further scaled by some power $\lambda$ of the singular values $\Sigma$
returned by the SVD solution.  In our experiments, we tune the choice
of $\lambda$ on the training data.  Trivially, if $\lambda=0$,
\model{nCCA} reduces to CCA.

Note that other mapping functions could also be used. We leave a more
extensive exploration of possible alternatives to further research,
since the details of how the vision-to-text conversion is conducted
are not crucial for the current study. As increasingly more effective
mapping methods are developed, we can easily plug them into our
architecture.


Through the selected cross-modal mapping function, any image can be
projected onto linguistic space, where the word (possibly of the
appropriate part of speech) corresponding to the nearest vector is
returned as a candidate label for the image (following standard
practice in distributional semantics, we measure proximity by the
\textit{cosine} measure).

\subsection{Decomposition}
\label{subsec:decomposition}

\newcite{Dinu:Baroni:2014} have recently proposed a general
decomposition framework that, given a distributional vector encoding a
phrase meaning and the syntactic structure of that phrase, decomposes
it into a set of vectors expected to express the semantics of the
words that composed the phrase.  In our setup, we are interested in a
decomposition function $f_{Dec}: \mathbb{R}^{d_{2}}\to
\mathbb{R}^{2{d_2}}$ which, given a visual vector projected onto the
linguistic space, assumes it represents the meaning of an
\emph{adjective-noun phrase}, and decomposes it into two vectors
corresponding to the adjective and noun constituents
$[w_{adj};w_{noun}]=f_{Dec}(w_{AN})$.  We take $f_{Dec}$ to be a
linear function and, following \newcite{Dinu:Baroni:2014}, we use as
training data vectors of adjective-noun bigrams directly extracted
from the corpus together with the concatenation of the corresponding
adjective and noun word vectors. We estimate $f_{Dec}$ by solving a
ridge regression problem minimizing the following
objective: $$\footnotesize ||[{W^{Tr}_{adj}};{W^{Tr}_{noun}}] -
F_{dec}W^{Tr}_{AN}||^{2}_{2} - ||\lambda F_{dec}||^{2}_{2} $$ where
$W^{Tr}_{adj}$, $W^{Tr}_{noun}$, $W^{Tr}_{AN}$ are the matrices
obtained by stacking the training data vectors. The $\lambda$
parameter is tuned through generalized cross-validation
\cite{Hastie:etal:2009}.

\subsection{Representational Spaces}
\label{sec:ss}

\paragraph{Linguistic Space}
We construct distributional vectors from text through the method
recently proposed by \newcite{Mikolov:etal:2013b}, to which we feed a
corpus of 2.8 billion words obtained by concatenating English
Wikipedia, ukWaC and BNC.\footnote{\url{http://wacky.sslmit.unibo.it},
  \url{http://www.natcorp.ox.ac.uk}} Specifically, we used the CBOW
algorithm, which induces vectors by predicting a target word given the
words surrounding it.  We construct vectors of 300 dimensions
considering a context window of 5 words to either side of the target,
setting the sub-sampling option to 1e-05 and the negative sampling
parameter to 5.\footnote{The parameters are tuned on the MEN word
  similarity dataset \cite{Bruni:etal:2014}.}

\paragraph{Visual Spaces}
Following standard practice, images are represented as bags of visual words
(BoVW) \cite{Sivic:2003}.\footnote{In future research, we might obtain a
performance boost simply by using the more advanced visual features
recently introduced by \newcite{Krizhevsky:etal:2012}.} Local low-level
image features are clustered into a set of \textit{visual words} that act
as higher-level descriptors.  In our case, we use PHOW-color image 
features, a variant of dense SIFT \cite{Bosch:etal:2007}, and
a visual vocabulary of 600 words.  Spatial information is preserved with a
two-level spatial pyramid representation~\cite{Lazebnik:2006}, achieving a
final dimensionality of 12,000. The entire pipeline is implemented using
the VLFeat library~\cite{Vedaldi:2010}, and its setup is identical to the
toolkit's \textit{basic recognition} sample
application.\footnote{\url{http://www.vlfeat.org/applications/apps.html}}
We apply Positive Pointwise Mutual Information \cite{Evert:2005} to the
BoVW counts, and reduce the resulting vectors to 300 dimensions using
SVD.

\subsection{Evaluation Dataset}
\label{subsec:eval-dataset}

For evaluation purposes, we use the dataset consisting of images
annotated with adjective-noun phrases introduced in
\newcite{Russakovsky:Fei-Fei:2010}, which pertains to 384
WordNet/ImageNet synsets with 25 images per synset. The images were
manually annotated with 25 attribute-denoting adjectives related to
texture, color, pattern and shape, respecting the constraints that a
color must cover a significant part of the target object, and all
other attributes must pertain to the object as a whole (as opposed to
parts). Table~\ref{tab:attributes} lists the 25 attributes and
Table~\ref{tab:eval_data} illustrates sample
annotations.\footnote{Although \textit{vegetation} is a noun, we have
  kept it in the evaluation set, treating it as an adjective.}

\begin{table}
\footnotesize
    \centering
    \begin{tabular}{l | l}
        Category &Attributes \\
        \hline
        Color &\emph{black, blue, brown, gray, green,} \\
              &\emph{orange, pink, red, violet, white, yellow} \\
        Pattern &\emph{spotted, striped} \\
        Shape &\emph{long, round, rectangular, square} \\
        Texture &\emph{furry, smooth, rough, shiny, metallic,} \\
                &\emph{vegetation, wooden, wet}
    \end{tabular}
    \caption{List of attributes in the evaluation dataset.} 
    \label{tab:attributes}
\end{table}

\begin{table}
    \footnotesize
    \begin{center}
        \begin{tabular}{P{1.1cm}ll}
            \textbf{Image} & \textbf{Attributes} & \textbf{Object}\\
            \hline\\[-2.1ex]
            \multirow{3}{*}{\includegraphics[scale=0.070]{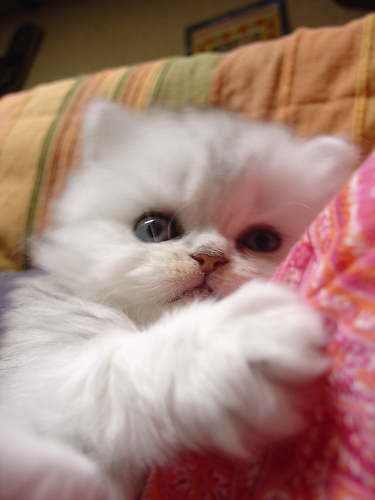}} & furry & cat \\
            & white & \\
            & smooth & \\[-1.9ex]\\
            \hline\\[-2.1ex]
            \multirow{3}{*}{\includegraphics[scale=0.065]{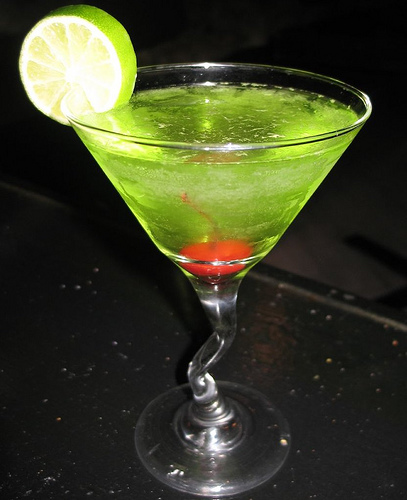}} & green & cocktail\\
            & shiny & \\
            & & 
        \end{tabular}
    \end{center}
    \caption{Sample annotations from the evaluation dataset.}
    \label{tab:eval_data}
\end{table}
 
In order to increase annotation quality, we only consider attributes
with full annotator consensus, for a total of 8,449 annotated images,
with 2.7 attributes per-image on average. Furthermore, to make the
linguistic annotation more natural and avoid sparsity problems, we
renamed excessively specific objects with a noun denoting a more
general category, following recent work on \emph{entry-level
  categories}~\cite{Ordonez:etal:2013}; e.g., \textit{colobus guereza}
was relabeled as \textit{monkey}. The final evaluation dataset
contains 203 distinct objects.

%
%
%

\begin{table}
\footnotesize
\centering
\begin{tabular}{@{}p{1cm}| p{0.5cm}  c  c| c  c  c@{}}
             & \multicolumn{3}{c}{Training}            & \multicolumn{3}{c}{Evaluation} \\
	     &  \#im. & \#attr.  & \#obj.&  \#im. & \#attr.  & \#obj.\\\hline
Exp.~1 &  10,749     &  97            & -         & \multicolumn{3}{c}{\textit{leave-one-attribute-out}}\\
Exp.~2 & 	23,000     &    - 	   & 750       &  8,449   & 25	     & 203\\
\end{tabular}
\caption{Summary of training and evaluation sets.}
\label{tab:trainingsets}
\end{table}

\section{Experiment 1: Zero-shot attribute learning}

\begin{figure*}
    \centering
    \includegraphics[scale=0.7]{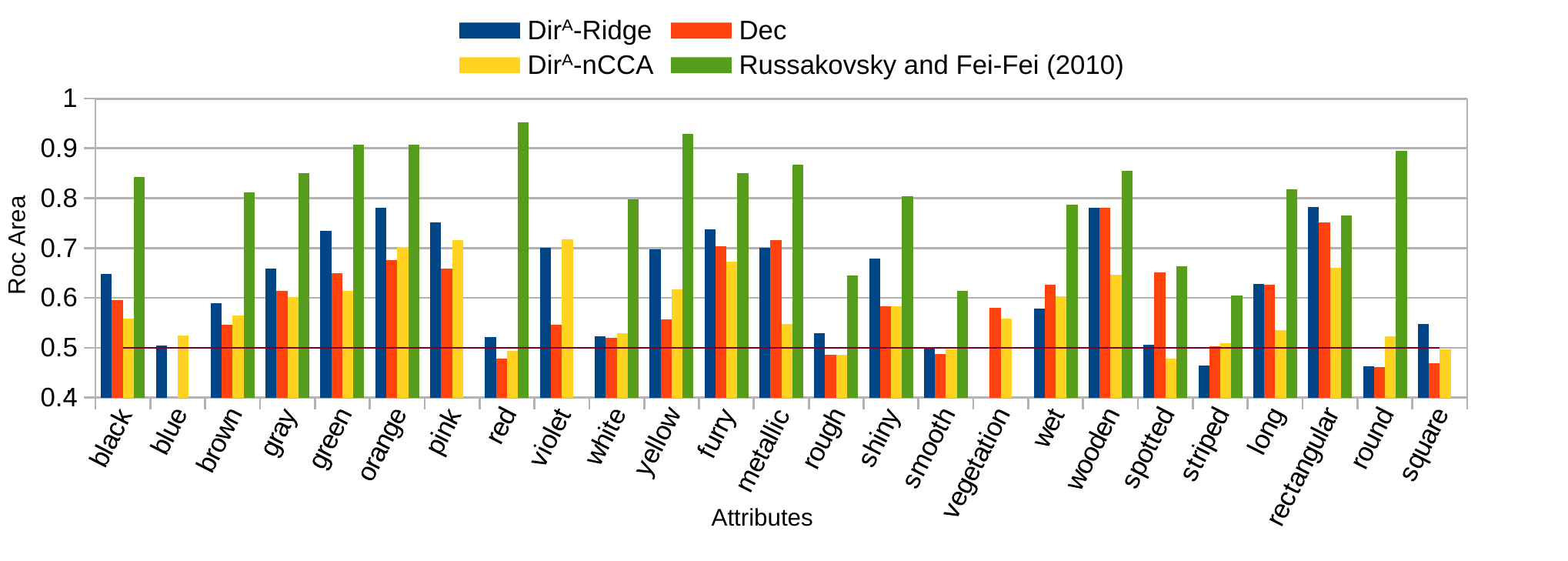}
    \caption{Performance of zero-shot attribute classification (as
      measured by AUC) compared to the supervised method of
      \protect\newcite{Russakovsky:Fei-Fei:2010}, where available. The
      dark-red horizontal line marks chance performance.}
    \label{fig:ROCchart}
\end{figure*} 

\label{sec:exp1}

In Section~\ref{sec:intro}, we showed that there is a significant
correlation between pairwise similarities of adjectives in a
language-based distributional semantic space and those of visual
feature vectors extracted from images labeled with the corresponding
attributes. In the first experiment, we test whether this
correspondence in attribute-adjective similarity structure across
modalities suffices to successfully apply zero-shot labeling. We learn
a cross-modal function from an annotated dataset and use it to label
images from an evaluation dataset with attributes outside the training
set. %
We will refer to this approach as
\textbf{\model{DiR\textsuperscript{A}}}, for \emph{\textbf{Di}rect
  \textbf{R}etrieval} using \emph{\textbf{A}ttribute} annotation. Note
that this is the first time that zero-shot techniques are used in the
attribute domain. In the present evaluation, we distinguish
\textbf{\model{DiR\textsuperscript{A}-Ridge}} and
\textbf{\model{DiR\textsuperscript{A}-\model{nCCA}}}, according to the
cross-modal function used to project from images to linguistic
representations (see Section \ref{subsec:mapping} above).

\subsection{Cross-modal training and evaluation} 
\label{sec:trmap1}
To gather sufficient data to train a cross-modal mapping function for
attributes/adjectives, we combine the publicly available datasets of
\newcite{Fahradi:etal:2009} and \newcite{Ferrari:Zisserman:2007} with
attributes and associated images extracted from
MIRFLICKR~\cite{Huiskes:Lew:2008}.\footnote{We filtered out attributes not
expressed by adjectives, such as \textit{wheel} or \textit{leg}.} The
resulting dataset contains 72 distinct attributes and 2,300 images.
Each image-attribute pair represents a training data point $(v,
w_{adj})$, where $v$ is the vector representation of the image, and
$w_{adj}$ is the linguistic vector of the attribute
(corresponding to an adjective). No information about the depicted
object is needed.

To further maximize the amount of training data points, we conduct a
leave-one-attribute-out evaluation, in which the cross-modal mapping
function is repeatedly learned on all 72 attributes from the training
set, as well as all but one attribute from the evaluation set (Section
\ref{subsec:eval-dataset}), and the associated images. This results in
$72+(25-1) = 96$ training attributes in total. On average, 45 images
per attribute are used. The performance is measured for the single
attribute that was excluded from training. A numerical summary of the
experiment setup is presented in the first row of Table
\ref{tab:trainingsets}.


\subsection{Results and discussion}


\newcite{Russakovsky:Fei-Fei:2010} trained separate SVM classifiers for
each attribute in the evaluation dataset in a cross-validation setting.
This fully supervised approach can be seen as an ambitious upper bound for
zero-shot learning, and we directly compare our performance to theirs using
their figure of merit, namely area under the ROC curve (AUC), which is
commonly used for binary classification problems.\footnote{Table
    \ref{tab:UNoun_Hit} reports hit@$k$ results for
\model{DiR\textsuperscript{A}}, which will be discussed below in the
context of Experiment 2.} A perfect classifier achieves an AUC of $1$,
whereas an AUC of $0.5$ indicates random guessing.  For purposes of AUC
computation, \model{DiR\textsuperscript{A}} is considered to label test
images with a given adjective if the linguistic-space distance between
their mapped representation and the adjective is below a certain threshold.
AUC measures the aggregated performance over all thresholds. To get a sense
of what AUC compares to in terms of precision and recall, the AUC of
\model{DiR\textsuperscript{A}} for \emph{furry} is 0.74, while the
precision is 71\% and the corresponding recall 14\%. For the more difficult
\emph{blue} case, AUC is at 0.5, precision and recall are 2\% and 55\%,
respectively.


The AUC results are presented in Figure~\ref{fig:ROCchart} (ignore red
bars for now). We observe first that, of the two mapping functions we
considered, \model{Ridge} (blue bars) clearly outperforms \model{nCCA}
(yellow bars). According to a series of paired permutation tests,
\model{Ridge} has a significantly larger AUC in 13/25 cases, 
\model{nCCA} in only 2. 
This is somewhat surprising given the better performance of \model{nCCA} in
the experiments of \newcite{Gong:etal:2014}. However, our setup is
quite different from theirs: They perform all retrieval tasks by
projecting the input visual and language data onto a common multimodal
space \textit{different} from both input spaces.  \model{nCCA} is a
well-suited algorithm for this. We aim instead at producing linguistic
annotations of images, which is most straightforwardly accomplished by
projecting visual representations onto linguistic
space. Regression-based learning (in our case, via \model{Ridge}) is a
more natural choice for this purpose. %

Coming now to a more general analysis of the results, as expected, and
analogously to the supervised setting,
\model{DiR\textsuperscript{A}-Ridge} performance varies across
attributes. Some achieve performance close to the supervised model
(e.g., \emph{rectangular} or \emph{wooden}) and, for 18 out of 25, the
performance is well above chance (bootstrap test). 
The exceptions are: \emph{blue, square, round, vegetation,
  smooth, spotted} and \emph{striped}. Interestingly, for the last 4
attributes in this list, \newcite{Russakovsky:Fei-Fei:2010} achieved
their lowest performance, attributing it to the
lower-quality of the corresponding image annotations.  Furthermore,
\newcite{Russakovsky:Fei-Fei:2010} excluded 5 attributes due to
insufficient training data. Of these, our performance for \emph{blue},
\emph{vegetation} and \emph{square} is not particularly encouraging,
but for \emph{violet} and \emph{pink} we achieve more than 0.7 AUC, at
the level of the supervised classifiers, suggesting that the proposed
method can complement the latter when annotated data are not
available.

For a different perspective on the performance of
\model{DiR\textsuperscript{A}}, we took several objects and queried
the model for their most common attribute, based on the average
attribute rank across all images of the object in the dataset.
Reassuringly, we learn that \emph{sunflowers} are on average
\emph{yellow} (mean rank 2.3), \emph{fields} are \emph{green} (4.4),
\emph{cabinets} are \emph{wooden} (4) and \emph{vans} \emph{metallic}
(6.6) (\emph{strawberries} are, suspiciously, \emph{blue}, 2.7).

Overall, this experiment shows that, just like object classification,
attribute classifiers benefit from knowledge transfer between the
visual and linguistic modalities, and zero-shot learning can achieve
reasonable performance on attributes and the corresponding
adjectives. This conclusion is based on the assumption
that \emph{per-image} annotations of attributes are available; in the
following section, we show how equal and even better performance can
be attained using data sets annotated with objects only, therefore
without any hand-coded attribute information.

\section{Experiment 2: Learning attributes from objects and visual phrases}

Having shown that reasonably accurate annotations of unseen attributes
can be obtained with zero-shot learning when a small amount of manual
annotation is available, we now proceed to test the intuition,
preliminarily supported by the data in Figure \ref{fig:bundle}, that,
since objects are bundles of attributes, attributes are implicitly
learned together with objects. We thus try to induce
attribute-denoting adjective labels by exploiting only
widely-available object-noun data.  At the same time, building on the
observation illustrated in Figure \ref{fig:visual_phrase} that
pictures of objects are pictures of visual phrases, we experiment with
a \textit{vector decomposition} model which treats images as composite
and derives adjective and noun annotations \emph{jointly}.  We compare
it with standard zero-shot learning using direct label retrieval as
well as against a number of challenging alternatives that exploit
gold-standard information about the depicted objects. The second row
of Table \ref{tab:trainingsets} gives a numerical summary of the setup
for this experiment.


\subsection{Cross-modal training} 
\label{sec:exp2_tr} 

We now assume object annotations only, in the form of training data 
$(v, w_{noun})$, where $v$ is the vector representation of an
image tagged with an object and $w_{noun}$ is the linguistic vector
of the corresponding noun. To ensure high imageability and
diversity, we use as training object labels those appearing in the
CIFAR-100 dataset~\cite{Krizhevsky:Hinton:2009}, combined with those
previously used in the work of \newcite{Fahradi:etal:2009}, as well as
the most frequent nouns in our corpus that also exist in ImageNet,
for a total of 750 objects-nouns.  For each object label, we include
at most 50 images from the corresponding ImageNet synset, resulting in
$\approx{}\!\!23,000$ training data points. Images containing objects
from the evaluation dataset are excluded, so that both adjective and
noun retrieval adhere to the zero-shot paradigm.

\subsection{Object-agnostic models} \label{sec:exp2_gen}

\paragraph{\model{DiR\textsuperscript{O}}} The \emph{\textbf{Di}rect
  \textbf{R}etrieval using \textbf{O}bject annotation} approach
projects an image onto the linguistic space and retrieves the nearest
adjectives as candidate attribute labels.  The only difference with
\model{DiR\textsuperscript{A}} (more precisely,
\model{DiR\textsuperscript{A}-Ridge}), the zero-shot approach we
tested above, is that the mapping function has been trained on
object-noun data only.

\paragraph{\model{Dec}} The \emph{\textbf{Dec}omposition} method uses
the $f_{Dec}$ function inspired by \newcite{Dinu:Baroni:2014} (see Section \ref{subsec:decomposition}),
 to associate
the image vector projected onto linguistic space to an adjective and a
noun. We train $f_{Dec}$ with about $\approx\!\!50,000$ training
instances, selected based on corpus frequency.  These data are further
balanced by not allowing more than 100 training samples for any
adjective or noun in order to prevent very frequent words such as
\emph{other} or \emph{new} from dominating the training data. No image
data are used, and there is no need for manual annotation, as the
adjective-noun tuples are automatically extracted from the corpus.

At test time, given an image to be labeled, we project its visual
representation onto the linguistic space and decompose the resulting
vector $w^{\prime}$ into two candidate adjective and noun vectors:
$[w_{adj}^{\prime};w_{noun}^{\prime}]=f_{Dec}(w^{\prime})$.  We then
search the linguistic space for adjectives and nouns whose vectors are
nearest to $w_{adj}^{\prime}$ and $w_{noun}^{\prime}$, respectively.

\subsection{Object-informed models} \label{sec:exp2_baselines}

A cross-modal function trained exclusively on object-noun data might
be able to capture only prototypical characteristics of an object, as
induced from text, independently of whether they are depicted in an
image.  Although the gold annotation of our dataset should already
penalize this image-independent labeling strategy (see
Section~\ref{subsec:eval-dataset}), we control for this behaviour by
comparing against three models that have access to the gold noun
annotations of the image and favor adjectives that are typical
modifiers of the nouns.


\paragraph{\model{LM}} We build a bigram \emph{\textbf{L}anguage
  \textbf{M}odel} by using the Berkeley LM toolkit
\cite{Pauls:Klein:2012}\footnote{\url{https://code.google.com/p/berkeleylm/}}
on the one-trillion-token Google Web1T
corpus\footnote{\url{https://catalog.ldc.upenn.edu/LDC2006T13}} and
smooth probabilities with the ``Stupid'' backoff
technique~\cite{Brants:etal:2007}.  Given an image with
object-\textbf{noun} annotation, we score all attributes-adjectives
based on the language-model-derived conditional probability
$p(adjective|\textbf{noun})$. All images of the same object produce
identical rankings. As an example, among the top attributes of
\emph{cocktail} we find \emph{heady}, \emph{creamy} and \emph{fruity}.

\paragraph{\model{vLM}} \model{LM} does not exploit visual information
about the image to be annotated. A natural way to enhance it is to
combine it with \model{DiR\textsuperscript{O}}, our cross-modal
mapping adjective retrieval method. In the
\emph{\textbf{v}isually-enriched \textbf{L}anguage \textbf{M}odel}, we
interpolate (using equal weights) the ranks produced by the two models.  
In the resulting combination, attributes that
are both linguistically sensible and likely to be present in the given
image should be ranked highest.  We expect this approach to be
challenging to beat. \newcite{Mackenzie:2014} recently introduced a
similar model in a supervised setting, where it improved over standard
attribute
classifiers.  

\paragraph{\model{SP}} The \emph{\textbf{S}electional
  \textbf{P}reference} model robustly captures semantic restrictions
imposed by a noun on the adjectives modifying it \cite{Erk:etal:2010}.
Concretely, for each \textbf{noun} denoting a target object, we
identify a set of adjectives $ADJ_{noun}$ that co-occur with it in a
modifier relation more that 20 times.  By averaging the linguistic
vectors of these adjectives, we obtain a vector
$w^{prototypical}_{\textbf{noun}}$, which should capture the semantics
of the \textit{prototypical} adjectives for that \textbf{noun}.
Adjectives that have higher similarity with this prototype vector are expected
to denote typical attributes of the corresponding noun and will be ranked as
more probable attributes. Similarly to \model{LM}, all images of the same
object produce identical rankings.  As an example, among the top attributes of
\emph{cocktail} we find \emph{fantastic}, \emph{delicious} and \emph{perfect}.

\subsection{Results} 
\label{sec:exp2_resnouns}

We evaluate the performance of the models on attribute-denoting
adjective retrieval, using a search space containing the top 5,000
most frequent adjectives in our corpus.  Tables \ref{tab:UNoun_Hit}
and \ref{tab:UNoun_Recall} present hit@$k$ and recall@$k$ results,
respectively ($k\in\{1,5,10,20,50,100\}$). \emph{Hit@$k$} measures the
percentage of images for which at least one gold attribute exists
among the top $k$ retrieved attributes. \emph{Recall@$k$} measures the
proportion of gold attributes retrieved among the top $k$, relative to
the total number of gold attributes for each image.\footnote{Due to
  the leave-one-attribute-out approach used to train and test
  \model{DiR\textsuperscript{A}} (see Section \ref{sec:exp1}), it is
  not possible to compute recall results for this model.}

\begin{table}
    \footnotesize
    \begin{tabular}{  l |r | r || r | r | r || r}
        & 	\model{LM} & \model{SP} & \model{vLM} & \model{DiR\textsuperscript{O}} & \model{Dec} & \model{DiR\textsuperscript{A}} \\\hline
      @1 &    2  & 0	  & 5	& 1 	& 10 & 7	\\	
      @5 & 	5  & 7  & 16	& 4	& 31 & 23	\\
      @10&	8  & 9  & 29	& 9	& 44 & 37	\\
      @20&	18 & 17 & 50	& 19	& 59 & 51	\\	
      @50 &	33 & 32 & 72	& 43	& 81 & 68	\\
      @100 &	56 & 55 & 82	& 67	& 89 & 77	\\
    \end{tabular}
    \caption{Percentage hit@$k$ attribute retrieval scores.}
    \label{tab:UNoun_Hit}
\end{table}

\begin{table}
    \centering
    \footnotesize
    \begin{tabular}{  l |r | r || r | r | r }
        & 	\model{LM} & \model{SP} & \model{vLM} & \model{DiR\textsuperscript{O}} & \model{Dec}  \\\hline
      @1 &    1  & 0	  & 2	& 0 	& 4	\\	
      @5 & 	2  & 3  & 7	& 2	& 15	\\
      @10&	3  & 5  & 15	& 4	& 23	\\
      @20&	9  & 10  & 30	& 9	& 35	\\	
      @50 &	20 & 20 & 49	& 22	& 59	\\
      @100 &	35 & 34 & 61	& 44	& 70	\\
    \end{tabular}
    \caption{Percentage recall@$k$ attribute retrieval scores.}
    \label{tab:UNoun_Recall}
\end{table}

First of all, we observe that \model{LM} and \model{SP} -- the two models that
have access to  gold object-noun annotation and are entirely language-based
-- although well above the random baseline ($k$/5,000), achieve rather low
performance. This confirms that to model our test set accurately, it is not
sufficient to predict typical attributes of the depicted objects.

The \model{DiR\textsuperscript{O}} method, which exploits visual
information, performs numerically similarly to the object-informed
models \model{LM} and \model{SP}, with better hit and recall at high
ranks.  Although worse than \model{DiR\textsuperscript{A}}, the
relatively high performance of \model{DiR\textsuperscript{O}} is a
promising result, suggesting object annotations together with
linguistic knowledge extracted in an unsupervised manner from large
corpora can replace, to some extent, manual attribute
annotations. 
However, \model{DiR\textsuperscript{O}} does not directly model any
\textit{semantic compatibility} constraints between the retrieved
adjectives and the object present in the image (see examples below).
Hence, the object-informed model \model{vLM}, which combines visual
information wit linguistic co-occurrence statistics, doubles the
performance of \model{DiR\textsuperscript{O}}, \model{LM} and
\model{SP}.


Our \model{Dec} model, which treats images as visual phrases and
jointly decouples their semantics, outperforms even \model{vLM} by a
large margin. %
It also outperforms \model{DiR\textsuperscript{A}}, the standard
zero-shot learning approach using attribute-adjective annotated data
(see also the attribute-by-attribute AUC comparison between
\model{Dec}, \model{DiR\textsuperscript{A}} and the fully-supervised
approach of Russakovsky and Fei-Fei in Figure
\ref{fig:ROCchart}). 

%

\begin{table}
    \centering
    \footnotesize
    \begin{tabular}{  l| r|  r || r}
       	&  \model{DiR\textsuperscript{O}}  & \model{Dec}  &  \model{DiR\textsuperscript{A}} \\\hline
        @1    &	1 & 2      &	 0              \\	
        @5    &	3 & 10	   &	 0              \\
        @10   &	5 & 14	   &	 1              \\
        @20   &	9 & 20	   &	 2               \\	
        @50   &	20 & 29	   &	 6               \\
        @100  &	33 & 41	   &	 12              \\
    \end{tabular}
    \caption{Percentage hit@$k$ noun retrieval scores.} 
    \label{tab:UNounNoun_Recall}
\end{table}

Interestingly, accounting for the phrasal nature of visual information
leads to substantial performance improvement in object recognition
through zero-shot learning (i.e., tagging images with the depicted
nouns) as well. Table \ref{tab:UNounNoun_Recall} provides the hit@$k$
results obtained with the \model{DiR\textsuperscript{O}} and
\model{Dec} methods for the \textbf{noun} retrieval task in a search
space of 10,000 most frequent nouns from our corpus.  Note that
\model{DiR\textsuperscript{O}} represents the label retrieval
technique that has been standardly used in conjunction with zero-shot
learning for objects: The cross-modal function is trained on images
annotated with nouns that denote the objects they depict, and it is
then used for noun label retrieval of unseen objects through a nearest
neighbor search of the mapped image representation (the
\model{DiR\textsuperscript{A}} column shows that zero-shot noun
retrieval using the mapping function trained on adjectives works very
poorly). \model{Dec} decomposes instead the mapped image
representation into two vectors denoting adjective and noun semantics,
respectively, and uses the latter to perform the nearest neighbor
search for a noun label.  Although not directly comparable, the
results of \model{Dec} reported here are in the same range of
state-of-the-art zero-shot learning models for object
recognition~\cite{Frome:etal:2013}.


 \begin{table}
\footnotesize

 
 \begin{tabular}{P{2.4cm}@{}P{1.1cm}@{}P{1.9cm}@{}l}\\
 \textbf{Image}& \textbf{Model} & \textbf{Top item} & \textbf{Top hit (Rank)}\\
 \hline\\ [-1.9ex]
\multirow{6}{*}{\parbox{2.3cm}{\includegraphics[scale=0.10]{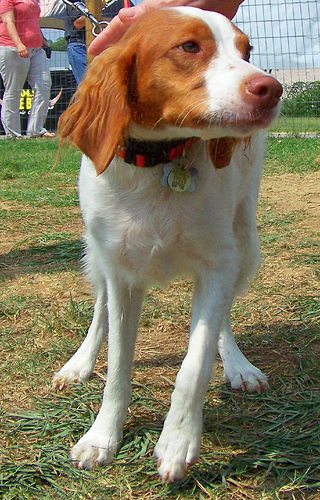}\\ A: white, brown\\ N: dog}} & \multirow{2}{*}{\model{Dec}}  & A: white  & white (1)\\
& & N: dog  & dog (1) \\
  \cline{2-4}
& \multirow{2}{*}{\model{DiR\textsuperscript{O}}} & A: animal  & white (27)\\
& & N: goat & dog (25)\\
  \cline{2-4}\\[-1.9ex]
& \multirow{1}{*}{\model{LM}} & A: stray & brown (74)\\
  \cline{2-4}\\[-1.9ex]
& \multirow{1}{*}{\model{vLM}} & A: pet & brown (17)\\
\\
  \hline\\ [-1.9ex]
\multirow{6}{*}{\parbox{2.3cm}{\includegraphics[scale=0.3]{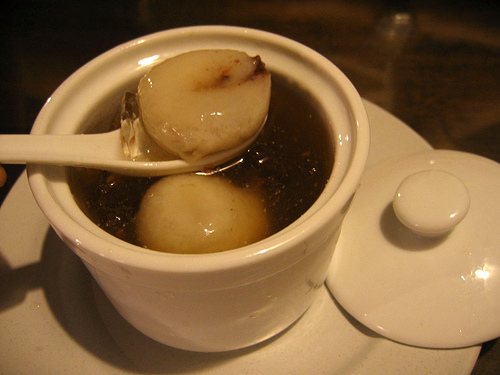} \\A: shiny, round\\ N: syrup}} & \multirow{2}{*}{\model{Dec}} & A: shiny  & shiny (1)\\
& & N: flan  & syrup (170)\\
  \cline{2-4}
& \multirow{2}{*}{\model{DiR\textsuperscript{O}}} & A: crunchy  & shiny (15)\\
& & N: ramekin  & syrup (113)\\
    \cline{2-4}\\[-1.9ex]
& \multirow{1}{*}{\model{LM}} & A: chocolate  & shiny (84)\\
  \cline{2-4}\\[-1.9ex]
& \multirow{1}{*}{\model{vLM}} & A: chocolate  & shiny (17)\\
\end{tabular}
\caption{Images with gold attribute-adjective and object-noun labels, and
    highest-ranked items for each model (\textbf{Top item}), as well as
    highest-ranked \textbf{correct} item and rank (\textbf{Top hit}). Noun
    results for  (\model{v})\model{LM} are omitted since these models have
    access to the gold noun label.}

\label{tab:qual_eval}
\end{table}


\paragraph{Annotation examples} Table
\ref{tab:qual_eval} presents some interesting patterns we observed in
the results.  The first example illustrates the case in which
conducting adjective and noun retrieval \textit{independently} results
in mixing information, which damages the
\model{DiR\textsuperscript{O}} approach: Adjectival and nominal
properties are not decoupled properly, since the \textit{animal}
property of the depicted dog is reflected in both the \emph{animal}
adjective and the \textit{goat} noun. At the same time, the
\textit{whiteness} of the object (an adjectival property) influences
noun selection, since \textit{goats} tend to be white. Instead,
\model{Dec} unpacks the visual semantics in an accurate and meaningful
way, producing correct attribute and noun annotations that form
acceptable phrases.  \model{LM} and \model{vLM} are negatively
affected by co-occurrence statistics and guess \emph{stray} and
\emph{pet} as adjectives, both typical but generic and abstract dog
properties.

In the next example, \model{DiR\textsuperscript{O}} predicts a reasonable
noun label (\emph{ramekin}), focusing on the
container rather than the liquid it contains. By ignoring the
relation between the adjective and the noun, the resulting adjective
annotation (\emph{crunchy}) is semantically incompatible with the noun label,
emphasizing the inability of this method to
account for semantic relations between attributes-adjectives and
object-nouns.  \model{Dec}, on the other hand, mistakenly annotates
the object as \emph{flan} instead of \emph{syrup}. However, having
captured the right general category of the object (``smooth gelatinous
items that reflect light''), it ranks a semantically appropriate and
correct attribute (\emph{shiny}) at the top. Finally, \model{LM} and \model{vLM} 
choose \emph{chocolate}, an attribute  semantically appropriate for \emph{syrup} but
irrelevant for the target image.


\paragraph{Semantic plausibility of phrases}The examples above suggest
that one fundamental way in which \model{Dec} improves over
\model{DiR\textsuperscript{O}} is by producing semantically coherent
adjective-noun combinations. More systematic evidence for this
conjecture is provided by a follow-up experiment on the linguistic
quality of the generated phrases. We randomly sampled 2 images for
each of the 203 objects in our data set. For each image, we let the
two models generate 9 descriptive phrases by combining their
respective top 3 adjective and noun predictions. From the resulting lists 
of 3,654 phrases, we picked the 200 most common ones for each model, 
with only 1/8 of these common phrases being shared by both. The selected phrases were
presented (in random order and concealing their origin) to two
linguistically-sophisticated annotators, who were asked to rate their 
degree of semantic plausibility on a 1-3 scale (the annotators
were not shown the corresponding images and had to evaluate
phrases purely on linguistic/semantic grounds). Since the two judges were
largely in agreement ($\rho\!\!=\!\!0.63$), we averaged their
ratings. The mean averaged plausibility score for
\model{DiR\textsuperscript{O}} phrases was 1.74 (s.d.: 0.76), for
\model{Dec} it was 2.48 (s.d.: 0.64), with the difference significant 
according to a Mann-Whitney test.
The two annotators agreed in assigning the
lowest score (``completely implausible'') to more than 1/3 of
the \model{DiR\textsuperscript{O}} phrases (74/200; e.g., \emph{tinned
  tostada, animal bird, hollow hyrax}), but they unanimously assigned
the lowest score to only 7/200 \model{Dec} phrases (e.g.,
\emph{cylindrical bed-sheet, sweet ramekin, wooden meat}). We thus
have solid quantitative support that the superiority of
\model{Dec} is partially due to how it learns to jointly
account for adjective and noun semantics, producing phrases that are
linguistically more meaningful.


\begin{figure}
\includegraphics[scale=0.6]{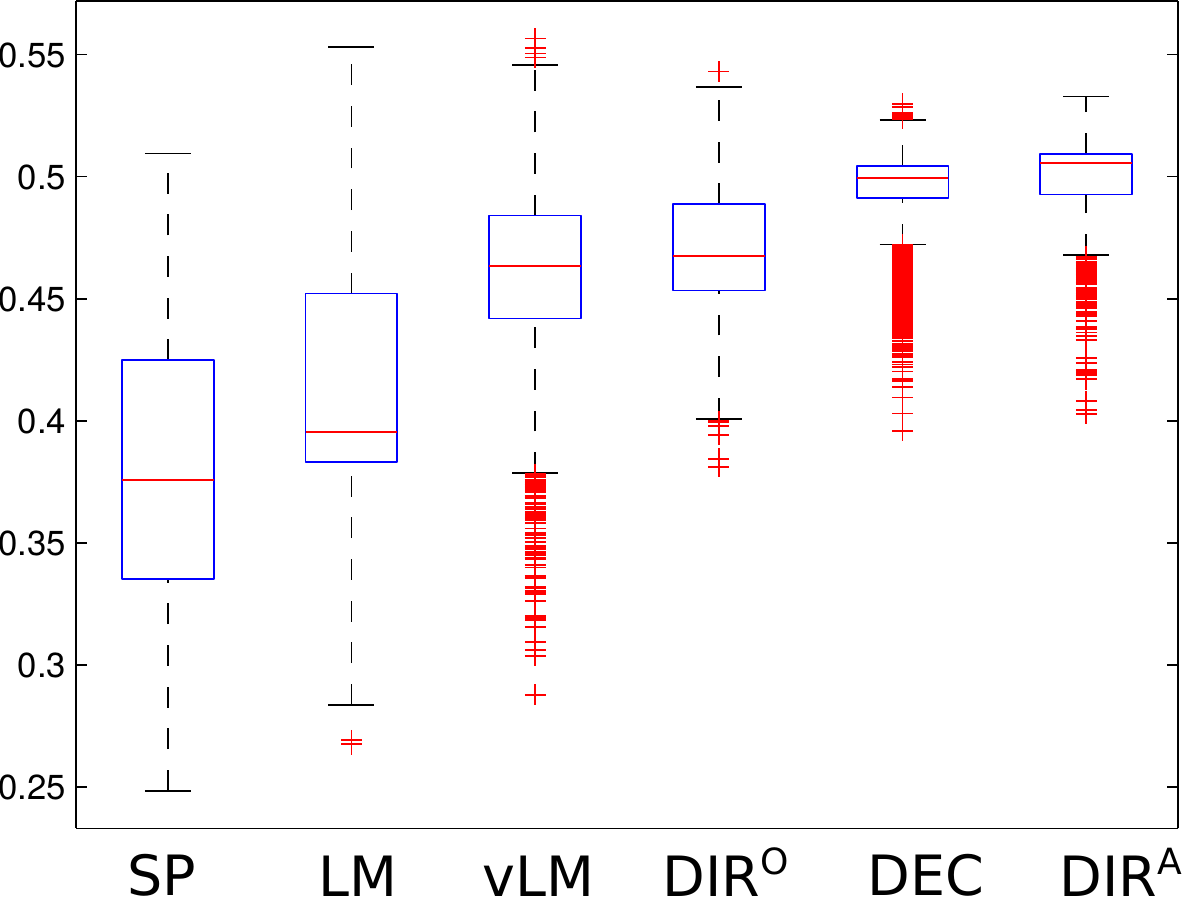}
\caption{Distributions of (per-image) concreteness scores across different models. Red line marks median values, box edges correspond to 1st and 3rd quartiles, the wiskers extend to the most extreme data points and outliers are plotted individually.}
\label{tab:concretness}
\end{figure}
 
\paragraph{Adjective concreteness}We can gain further insight into the
nature of the adjectives chosen by the models by considering the fact
that phrases that are meant to describe an object in a picture should
mostly contain concrete adjectives, and thus the degree of
concreteness of the adjectives produced by a model is an indirect
measure of its quality. Following \newcite{Hill:Korhonen:2014}, we
define the concreteness of an adjective as the average concreteness
score of the nouns it modifies in our text corpus. Noun concreteness
scores are taken, in turn, from \newcite{Turney:etal:2011}. For each
test image and model, we obtain a concreteness score by averaging the
concreteness of the top 5 adjectives that the model selected for the
image. Figure~\ref{tab:concretness} reports the distributions of the
resulting scores across models. We confirm that the purely
language-based models (\model{LM}, \model{SP}) are producing generic
abstract adjectives that are not appropriate to describe images (e.g.,
\textit{cryptographic} key, \textit{homemade} bread, \textit{Greek}
salad, \textit{beaten} yolk). The image-informed \model{vLM} and
\model{DiR\textsuperscript{O}} models produce considerably more
concrete adjectives. Not surprisingly, \model{DiR\textsuperscript{A}},
that was directly trained on concrete adjectives, produces the most
concrete ones. Importantly, \model{Dec}, despite being based on
a cross-modal function that was not explicitly exposed to adjectives,
produced adjectives that are approaching the concreteness level of
those of \model{DiR\textsuperscript{A}} (both differences between \model{Dec} and
\model{DiR\textsuperscript{O}}, \model{Dec} and
\model{DiR\textsuperscript{A}} are significant as by paired
Mann-Whitney tests).

\section{Using \model{Dec} for attribute-based object classification}
\label{sec:exp3}

As discussed in the introduction, attributes can effectively be used
for attribute-based object classification.  In this section, we show
that classifiers trained on attribute representations created with
\model{Dec} -- which does not require any attribute-annotated training
data nor training a battery of attribute classifiers -- outperform
(and are complementary to) standard BoVW features.


We use a subset of the Pascal VOC 2008
dataset.\footnote{\url{http://pascallin.ecs.soton.ac.uk/challenges/VOC/voc2008/}}
Specifically, following \newcite{Fahradi:etal:2009}, we use the original
VOC training set for training/validation, and the VOC validation set for
testing.
One-vs-all linear-SVM classifiers are trained for all VOC objects, using 3
alternative image representations.

First, we train directly on BoVW features (\textbf{\model{PHOW}}, see
Section \ref{sec:ss}), as in the classic object recognition pipeline.
We compare \model{PHOW} to an attribute-centric approach with
attribute labels automatically generated by \textbf{\model{Dec}}. All
VOC images are projected onto the linguistic space using the
cross-modal mapping function trained with object-noun data only (see
Section~\ref{sec:exp2_tr}), from which we further removed all images
depicting a VOC object.  Each image projection is then decomposed
through \model{Dec} into two vectors representing adjective and noun
information.  The final attribute-centric vector representing an image
is created by recording the cosine similarities of the
\model{Dec}-generated adjective vector with all the adjectives in our
linguistic space. Informally, this representation can be thought of as
a vector of weights describing the appropriateness of each adjective
as an annotation for the image.\footnote{Given that the resulting
  representations are very dense, we sparsify them by setting to zeros
  all adjective dimensions with cosine below the global mean cosine
  value.} This is comparable to standard attribute-based
classification \cite{Fahradi:etal:2009}, in which images are
represented as distributions over attributes estimated with a set of
\emph{ad hoc} supervised attribute-specific classifiers. Table
\ref{tab:VOC_example} show examples of top attributes automatically
assigned by \model{Dec}. While not nearly as accurate as manual
annotation, many attributes are relevant to the objects, both as
specifically depicted in the image (the aeroplane is \emph{wet}), but
also more prototypically (aeroplanes are \textit{cylindrical} in
general).
\begin{table}
    \footnotesize
    \begin{center}
        \begin{tabular}{P{2.7cm}@{}P{1.7cm}@{}P{2cm}}
            \textbf{Image} & \textbf{Object} & \textbf{Predicted Attributes}\\
            \hline \\[-1.9ex]
            \multirow{4}{*}{\includegraphics[scale=0.17]{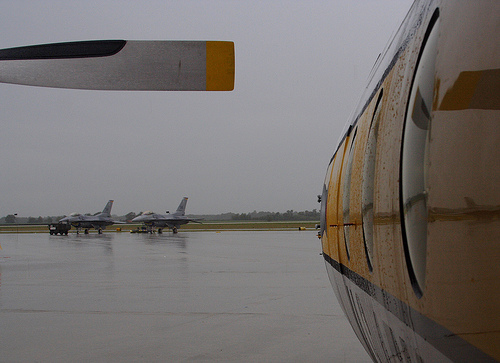}} & aeroplane & thick, wet, dry, cylindrical, motionless, translucent \\\\[-1.9ex]
            \hline \\[-1.9ex]
            \multirow{4}{*}{\includegraphics[scale=0.128]{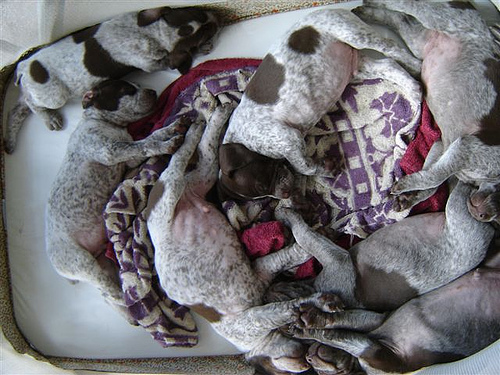}} & dog & cuddly, wild, cute, furry, white, coloured\\
\\
\end{tabular}
    \end{center}
    \caption{Two VOC images with  some top attributes  assigned by \model{Dec}: these attributes, together with their cosine similarities to the mapped image vectors, serve as attribute-centric representations.}
    \label{tab:VOC_example}
\end{table}

We also perform feature-level fusion (\textbf{\model{Fused}}) by
\textit{concatenating} the \model{PHOW} and \model{Dec} features, and
reducing the resulting vector to 100 dimensions with SVD
\cite{Bruni:etal:2014} (SVD dimensionality determined by
cross-validation on the training set).

\subsection{Results}
\begin{figure}[t]
  \includegraphics[scale=0.40]{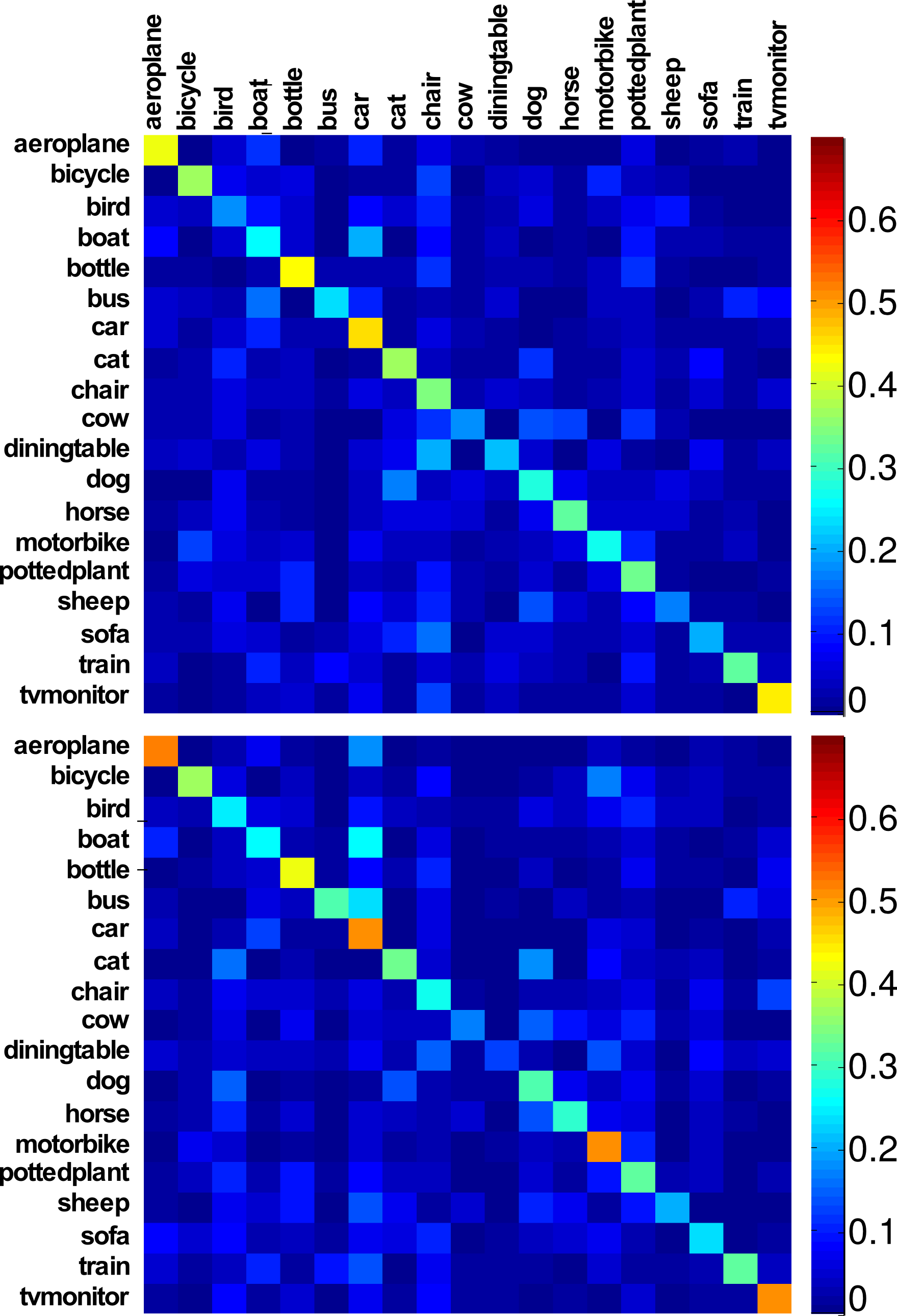}
  \caption{Confusion matrices for \model{PHOW} (\textbf{top}) and
    \model{Dec} (\textbf{bottom}). Warmer-color cells correspond to
    higher proportions of images with gold row label tagged by
    an algorithm with the column label (e.g., the first cells show
    that \model{Dec} tags a larger proportion of aeroplanes
    correctly).}
\label{fig:VOC}
\end{figure}

There is an improvement over PHOW visual features when using
\model{Dec}-based attribute vectors, with accuracy raising from
30.49\% to 32.76\%. The confusion matrices in Figure \ref{fig:VOC}
show that \model{PHOW} and \model{Dec} do not only differ in
quantitative performance, but make different kinds of errors, in part pointing at %
the different modalities the two models tap into. \model{PHOW}, for
example, tends to confuse \emph{cats} with \emph{sofas}, probably
because the former are often pictured lying on the
latter. \model{Dec}, on the other hand, tends to confuse \emph{chairs}
with \emph{TV monitors}, partially misguided by the taxonomic
information encoded in language (both are pieces of furniture). %
Indeed, the combined \model{Fused} approach outperforms both
representations by a large margin (35.81\%), confirming that
the linguistically-enriched information brought by \model{Dec} is to a
certain extent complementary to the lower-level visual evidence
directly exploited by PHOW. Overall, the performance of our system is quite close to the one obtained by
\newcite{Fahradi:etal:2009} with ensembles of supervised attribute
classifiers trained on manually annotated data (the most comparable
accuracy from their Table 1 is at 34.3\%).\footnote{Farhadi and
  colleagues reduce the bias for the \emph{people} category by
  reporting mean per-class accuracy; we directly excluded
  \emph{people} from our version of the data set.}

\section{Conclusion}
We extended zero-shot image labeling beyond objects, showing that it
is possible to tag images with attribute-denoting adjectives that
were not seen during training. For some attributes, performance was
comparable to that of per-attribute supervised classifiers. %
We further showed that attributes are \textit{implicitly} induced when
learning to map visual vectors of objects to their linguistic
realizations as nouns, and that improvements in both %
attribute \emph{and} noun retrieval are attained by treating images as
\emph{visual phrases}, whose linguistic representations must be
decomposed into a coherent word sequence. %
The resulting model outperformed a set of strong rivals. While the
performance of the zero-shot decompositional approach in the
adjective-noun phrase labeling alone might still be low for
practical applications, this model can still produce
attribute-based representations that significantly improve performance
in a supervised object recognition task, when combined with standard
visual features.

By mapping attributes and objects to phrases in a linguistic space, we
are also likely to produce more \emph{natural} descriptions than those
currently used in computer vision (\emph{fluffy kittens} rather than
\emph{2-boxy tables}). In future work, we want to delve more into the
linguistic and pragmatic naturalness of attributes: Can we predict not
just which attributes of a depicted object are true, but which are
more \emph{salient} and thus more likely to be mentioned (\emph{red
  car} over \emph{metal car})? Can we pick the most appropriate
adjective to denote an attribute given the object in the picture
(\emph{moist}, rather than \emph{damp lips})? We should also address
attribute dependencies: by ignoring
them, we currently get undesired results, such as the aeroplane in
Table \ref{tab:VOC_example} being tagged as both \emph{wet} and
\emph{dry}. %
More ambitiously, inspired by \newcite{Karpathy:etal:2014}, we plan
to associate image \textit{fragments} with phrases of arbitrary
syntactic structures (e.g., \textit{PP}s for backgrounds, a \textit{VP}s for
main events), paving the way to full-fledged caption generation.





\section*{Acknowledgments}
We thank the TACL reviewers for their feedback.
We were supported by ERC 2011 Starting
Independent Research Grant n.~283554 (COMPOSES).

\bibliographystyle{acl}
\bibliography{../../marco,../../elia,../../angeliki}
\end{document}